\documentclass{ifacconf} 
\usepackage{url}
\usepackage{amsmath,amssymb}
\usepackage{algorithm}
\usepackage{graphicx}
\usepackage{color}
\usepackage{xspace}
\usepackage{array}
\usepackage{natbib}

\usepackage{color}

\newcommand\ie{i.e.\xspace}
\newcommand\eg{e.g.\xspace}

\newcommand\myd{\text{d}}
\newcommand\E{\mathbb{E}}

\newcommand{\n}[1]{\mathbf{#1}}
\newcommand{\x}{\mathbf{x}}
\newcommand{\y}{\mathbf{y}}
\newcommand{\xu}{{\mathbf{x}}}
\newcommand{\T}{T} 

\newcommand{\btheta}{\boldsymbol\theta}

\newcommand{\kernelPGAS}[1][\btheta]{P_{#1}^N}

\DeclareMathOperator*{\argmax}{arg\,max}

\newcommand\ffun{f}
\newcommand\gfun{g}

\newcommand{\bmu}{\boldsymbol\mu}
\newcommand\Kt{\widetilde{\n{K}}} 
\newcommand\K{\n{K}}

\newcommand\gpssm{GP-SSM\xspace}

\begin{document}
\begin{frontmatter}  

\title{Identification of Gaussian Process State-Space Models with Particle Stochastic Approximation EM}


\author[First]{Roger Frigola} 
\author[Second]{Fredrik Lindsten} 
\author[Second,Third]{Thomas B. Sch\"{o}n}
\author[First]{Carl E. Rasmussen}

\address[First]{Dept. of Engineering, University of Cambridge, United Kingdom.}     
\address[Second]{Div. of Automatic Control, Link\"{o}ping University, Sweden.}
\address[Third]{Dept. of Information Technology, Uppsala University, Sweden.}


\maketitle

\begin{abstract}
Gaussian process state-space models (GP-SSMs) are a very flexible family of models of nonlinear dynamical systems.
They comprise a Bayesian nonparametric representation of the dynamics of the system and additional (hyper-)parameters
governing the properties of this nonparametric representation.
The Bayesian formalism enables systematic reasoning about the uncertainty in the system dynamics.
We present an approach to maximum likelihood identification of the parameters in GP-SSMs, while retaining the
full nonparametric description of the dynamics. The method is
based on a stochastic approximation version of the EM algorithm that employs recent developments in particle Markov chain Monte Carlo
for efficient identification. 


\end{abstract}

\end{frontmatter} 

\section{Introduction}

Inspired by recent developments in robotics and machine learning, we aim at constructing models of
nonlinear dynamical systems capable of quantifying the uncertainty in their predictions. To do so,
we use the Bayesian system identification formalism whereby degrees of uncertainty and belief
are represented using probability distributions \citep{Peterka1981}. Our goal is to identify models
that provide error-bars to any prediction. If the data is informative and the system identification
unambiguous, the model will report high confidence and error-bars will be narrow. On the other hand,
if predictions are made in operating regimes that were not present in the data used for system
identification, we expect the error-bars to be larger.

Nonlinear state-space models are a very general and widely used class of dynamical system models. They allow for modeling of systems based on observed input-output data through the use of a latent (unobserved) variable, the \emph{state} $\x_t \in \mathcal{X} \triangleq \mathbb{R}^{n_x}$. A discrete-time state-space model (SSM) can be described by
\begin{subequations}
  \label{eq:ssm}
  \begin{align}
    \label{eq:ssma}
    \x_{t+1} &= f(\x_{t},\n{u}_{t}) + \n{v}_{t}, \\
    \label{eq:ssmb}
    \n{y}_{t} &= g(\x_{t},\n{u}_{t}) + \n{e}_{t},
  \end{align}
\end{subequations}
where $\y_t$ represents the output signal, $\n{u}_t$ is the input signal, and $\n{v}_{t}$ and $\n{e}_{t}$ denote i.i.d. noises. The state transition dynamics is described by the nonlinear function $\ffun$ whereas~$\gfun$ links the output data at a given time to the latent state and input at that same time. For convenience, in the following we will not explicitly represent the inputs in our formulation. When available, inputs can be straightforwardly added as additional arguments to the functions $\ffun$ and~$\gfun$.

A common approach to system identification with nonlinear state-space models consists in defining a parametric form for the functions $\ffun$ and~$\gfun$ and finding the value of the parameters that minimizes a cost function, e.g. the negative likelihood. Those parametric functions are typically based on detailed prior knowledge about the system, such as the equations of motion of an aircraft, or belong to a class of parameterized generic function approximators, e.g. artificial neural networks (ANNs). In the following, it will be assumed that no detailed prior knowledge of the system is available to create a parametric model that adequately captures the complexity of the dynamical system. As a consequence, we will turn to generic function approximators. Parametrized nonlinear functions such as radial basis functions or other ANNs suffer from both theoretical and practical problems. For instance, a practitioner needs to select a parametric structure for the model, such as the number of layers and the number of neurons per layer in a neural network, which are difficult to choose when little is known about the system at hand. On a theoretical level, fixing the number of parameters effectively bounds the complexity of the functions that can be fitted to the data \citep{Gha12}. In order to palliate those problems, we will use Gaussian processes \citep{RasWil06} which provide a practical framework for Bayesian nonparametric nonlinear system identification.

Gaussian processes (GPs) can be used to identify nonlinear state-space models by placing
GP priors on the unknown functions. This gives rise to the Gaussian Process
State-Space Model (GP-SSM) \citep{TurDeiRas10,FriLinSchRas13} which will be introduced in
Section~\ref{sec:gpssm}. The \gpssm is a nonparametric model, though, the GP is
in general governed by a (typically) small number of hyper-parameters, effectively rendering the model semiparametric.
In this work, the hyper-parameters of the model will be estimated by maximum likelihood, while retaining the full nonparametric
richness of the system model. This is accomplished by analytically marginalizing out the nonparametric
part of the model and using the particle stochastic approximation EM (PSAEM) algorithm by \citet{Lin13} for estimating the parameters.


Prior work on GP-SSMs includes \citep{TurDeiRas10}, which presented an approach to maximum likelihood estimation in GP-SSMs based on analytical approximations and the parameterization of GPs with a pseudo data set. \citet{ko2011} proposed an algorithm to learn (i.e. identify) GP-SSMs based on observed data which also used weak labels of the unobserved state trajectory. \citet{FriLinSchRas13} proposed the use of particle Markov chain Monte Carlo to provide a fully Bayesian solution to the identification of GP-SSMs that did not need a pseudo data set or weak labels about unobserved states. However, the fully Bayesian solution requires priors on the model parameters which are unnecessary when seeking a maximum likelihood solution. Approaches for filtering and smoothing using already identified GP-SSMs have also been developed \citep{DeisenrothTurner2011,DeisenrothMohamed2012}.


\section{Gaussian Process State-Space Models}
\label{sec:gpssm}

\subsection{Gaussian Processes}

Whenever there is an unknown function, GPs allow us to perform Bayesian inference directly in the space of functions rather than having to define a parameterized family of functions and perform inference in its parameter space. GPs can be used as priors over functions that encode vague assumptions such as smoothness or stationarity. Those assumptions are often less restrictive than postulating a parametric family of functions.

Formally, a GP is defined as a collection of random variables, any finite number of which have a joint Gaussian distribution. A GP $f(\x) \in \mathbb{R}$ can be written as
\begin{equation}
	f(\x) \sim \mathcal{GP}\big(m(\x), k(\x,\x')\big),
\end{equation}
where the mean function $m(\x)$ and the covariance function $k(\x,\x')$ are defined as
\begin{subequations}\label{eq:meancovfun}
\begin{align}
m(\x) & = \mathbb{E}[f(\x)], \\
k(\x,\x') & = \mathbb{E}[(f(\x)-m(\x))(f(\x')-m(\x'))].
\end{align}
\end{subequations}

A finite number of variables from a Gaussian process follow a jointly Gaussian distribution 
\begin{equation}
\begin{bmatrix}
f(\x_1) \\
f(\x_2) \\
\vdots  \\
\end{bmatrix} \sim \mathcal{N}\left(
\begin{bmatrix}
m(\x_1) \\
m(\x_2) \\
\vdots  \\
\end{bmatrix},
\begin{bmatrix}
k(\x_1,\x_1) & k(\x_1,\x_2) &   \\
k(\x_2,\x_1) & k(\x_2,\x_2) &\\
 & &  \ddots  \\
\end{bmatrix}\right).
\end{equation}
We refer the reader to \citep{RasWil06} for a thorough exposition of GPs.


\subsection{Gaussian Process State-Space Models}

In this article we will focus on problems where there is very little information about the nature of the state transition function $f(\x_t)$ and a GP is used to model it. However, we will consider that more information is available about $g(\x_t)$ and hence it will be modeled by a parametric function. This is reasonable in many cases where the mapping from states to observations is known, at least up to some parameters.

The generative probabilistic model for the GP-SSM is fully specified by
\begin{subequations}\label{eq:generativemodel}
\begin{align}
\n{f}_{t+1} \mid \x_{t} &\sim \mathcal{GP}\big(m_{\btheta_\x}(\x_{t}), k_{\btheta_\x}(\x_{t}, \x_{t}^\prime)\big),  \label{eq:gpprior} \\ 
\x_{t+1} \mid \n{f}_{t+1} &\sim \mathcal{N}(\x_{t+1} \mid \n{f}_{t+1}, \n{Q}), \label{eq:processnoise} \\
\y_t \mid \x_t &\sim p(\y_t \mid \x_t , \btheta_\y), \label{eq:likelihood}
\end{align}
\end{subequations}
where $\n{f}_{t+1} = f(\x_t)$ is the value taken by the state $\x_{t+1}$ after passing through the transition function, but before the application of process noise $\mathbf{v}_{t+1}$. The Gaussian process in \eqref{eq:gpprior} describes the prior distribution over the transition function. The GP is fully specified by its mean function~$m_{\btheta_\x}(\x)$ and its covariance function~$k_{\btheta_\x}(\x_{t}, \x_{t}^\prime)$, which are parameterized by the vector of hyper-parameters $\btheta_\x$. Equation \eqref{eq:processnoise} describes the addition of process noise following a zero-mean Gaussian distribution of covariance~$\n{Q}$. We will place no restrictions on the likelihood distribution~\eqref{eq:likelihood} which will be parameterized by a finite-dimensional vector~$\btheta_\y$. For notational convenience we group all the \mbox{(hyper-)}parameters into a single vector~$\btheta = \{\btheta_{\x},\btheta_{\y},\n{Q}\}$.


\section{Maximum Likelihood in the GP-SSM}
\label{sec:maxlik}

Maximum likelihood (ML) is a widely used frequentist estimator of the parameters in a statistical
model. The ML estimator $\widehat{\btheta}^{\text{ML}}$ is defined as the value of the parameters
that makes the available observations $\y_{0:T}$ as likely as possible according ot,
\begin{equation}
  \widehat{\btheta}^{\text{ML}} = \argmax_{\btheta} \ p(\y_{0:T} \mid \btheta).
\end{equation} 
The GP-SSM has two types of latent variables that need to be marginalized (integrated out) in order to compute the likelihood
\begin{align}
  \nonumber
  &p(\y_{0:T} \mid \btheta) = \int p(\y_{0:T}, \x_{0:T}, \n{f}_{1:T} \mid \btheta) \ \text{d}\x_{0:T} \, \text{d}\n{f}_{1:T} \\
  \label{eq:marginallik0}
  &= \int p(\y_{0:T} \mid \x_{0:T}, \btheta) \left( \int p(\x_{0:T}, \n{f}_{1:T} \mid \btheta)\  \text{d}\n{f}_{1:T} \right) \text{d}\x_{0:T}.
\end{align}
Following results from \citet{FriLinSchRas13}, the latent variables $\n{f}_{1:T}$ can be marginalized analytically. This is equivalent to integrating out the uncertainty in the unknown function $\ffun$ and working directly with a prior over the state trajectories $p(\x_{0:\T} \mid \btheta)$ that encodes the assumptions (\eg smoothness) of $\ffun$ specified in~\eqref{eq:gpprior}. The prior over trajectories can be factorized as
\begin{align}
  \label{eq:fmarginalised}
  p(\x_{0:\T} \mid \btheta) = p(\x_0 \mid \btheta) \prod_{t=1}^T  p(\x_t \mid \btheta,\x_{0:t-1}).
\end{align}
Using standard expressions for GP prediction, the one-step predictive density is given by
\begin{subequations}\label{eq:allsequential}
\begin{align}
 p(\x_t \mid \btheta,\x_{0:t-1}) =
  \mathcal{N}\big(\x_t \mid \bmu_t(\x_{0:t-1}), \n{\Sigma}_t(\x_{0:t-1})\big) ,
\end{align}
where 
  \begin{align}
    \label{eq:model:mean_cov_a}
    \bmu_t(\x_{0:t-1}) &= \n{m}_{t-1} + \K_{t-1,0:t-2} \Kt^{-1}_{0:t-2} \ (\x_{1:t-1} - \n{m}_{0:t-2}), \\
    \label{eq:model:mean_cov_b}
    \n{\Sigma}_t(\x_{0:t-1}) &= \Kt_{t-1} - \K_{t-1,0:t-2} \Kt^{-1}_{0:t-2} \K_{t-1,0:t-2}^\top,
  \end{align}
\end{subequations}
for $t\geq2$ and $\bmu_1(\x_0) = \n{m}_0$, $\n{\Sigma}_1(\x_0) = \Kt_0$. Here we have defined the mean vector
  $\n{m}_{0:t-1} \triangleq
  \begin{bmatrix}
    m(\xu_0)^\top & \dots & m(\xu_{t-1})^\top
  \end{bmatrix}^\top$
and the $(n_x t) \times (n_x t)$ positive definite matrix $\K_{0:t-1}$ with block entries
$[\K_{0:t-1}]_{i,j} = k(\xu_{i-1}, \xu_{j-1})$. 
These matrices use two sets of indices, as in $\K_{t-1,0:t-2}$, to refer to the off-diagonal blocks of $\K_{0:t-1}$.
We also define $\Kt_{0:t-1} = \K_{0:t-1} + \n{I}_t \otimes \n{Q}$, where $\otimes$ denotes the Kronecker product.

Using \eqref{eq:fmarginalised} we can thus write the likelihood \eqref{eq:marginallik0} as
\begin{align}
  \label{eq:marginallik}
  p(\y_{0:T} \mid \btheta)= \int p(\y_{0:T} \mid \x_{0:T}, \btheta) p(\x_{0:T}\mid \btheta) \ \text{d} \x_{0:T}.
\end{align}
The integration with respect to $\x_{0:\T}$, however, is not analytically tractable. This difficulty will be addressed
in the subsequent section.

A GP-SSM can be seen as a hierarchical probabilistic model which describes a prior over the latent
state trajectories $p(\x_{0:\T} \mid \btheta_\x,\n{Q})$ and links this prior with the observed data
via the likelihood $p(\y_t \mid \x_t , \btheta_\y)$. Direct application of maximum likelihood on $p(\y_t \mid \x_t , \btheta_\y)$ to obtain estimates of the state trajectory and likelihood parameters would invariably result in over-fitting. However, by introducing a prior on the state trajectories\footnote{A prior over the state trajectories is not an exclusive feature of GP-SSMs. Linear-Gaussian state-space models, for instance, also describe a prior distribution over state trajectories: $p(\x_{1:\T} \mid \n{A},\n{B},\n{Q},\x_0)$.} and marginalizing them as in~\eqref{eq:marginallik}, we obtain the so-called marginal likelihood. Maximization of the marginal likelihood with respect to the parameters results in a procedure known as type II maximum likelihood or empirical Bayes \citep{Bishop2006}. Empirical Bayes reduces the risk of over-fitting since it automatically incorporates a trade-off between model fit and model complexity, a property often known as Bayesian Occam's razor \citep{Gha12}.


\section{Particle Stochastic Approximation EM}
\label{sec:psaem}

As pointed out above, direct evaluation of the likelihood \eqref{eq:marginallik} is not possible for a \gpssm.
However, by viewing the latent states $\x_{0:\T}$ as missing data, we are able to evaluate the \emph{complete data}
log-likelihood
\begin{equation}
\label{eq:completedataloglik}
	\log p(\y_{0:T}, \x_{0:T} \mid \btheta) = \log p(\y_{0:T}\mid \x_{0:T} , \btheta) + \log p(\x_{0:T} \mid \btheta),
\end{equation}
by using~\eqref{eq:likelihood} and~\eqref{eq:allsequential}. 
We therefore turn to the Expectation Maximization (EM) algorithm \citep{EM77}.  The EM algorithm uses \eqref{eq:completedataloglik}
 to construct a surrogate
cost function for the ML problem, defined as
\begin{align}
  \nonumber
  Q({}&\btheta, \btheta^\prime) = \E_{\btheta^\prime}[ \log p(\y_{0:T}, \x_{0:T} \mid \btheta) \mid y_{0:\T}]\\
  \label{eq:Qdef}
  &=  \int \log p(\y_{0:T}, \x_{0:T} \mid \btheta) p(\x_{0:T} \mid \y_{0:T}, \btheta^\prime) \myd\x_{0:\T}.
\end{align}
It is an iterative procedure that maximizes \eqref{eq:marginallik} by iterating two steps, expectation~(E) and maximization~(M),
\begin{itemize}
\item[(E)] Compute $Q(\btheta, \btheta_{k-1})$.
\item[(M)] Compute $\btheta_k = \argmax_{\btheta} Q(\btheta, \btheta_{k-1})$.
\end{itemize}
The resulting sequence $\{\btheta_k\}_{k\geq 0}$ will, under weak assumptions, converge
to a stationary point of the likelihood $p(\y_{0:\T} \mid \btheta)$.

To implement the above procedure we need to compute the integral in \eqref{eq:Qdef}, which in general is not computationally tractable
for a GP-SSM. To deal with this difficulty, 
we employ a Monte-Carlo-based implementation of the EM algorithm, referred to as PSAEM \citep{Lin13}.
This procedure is a combination of stochastic approximation EM (SAEM) \citep{DeyLavMou99} and
particle Markov chain Monte Carlo (PMCMC) \citep{Andrieu2010,LinJorSch12}.
As illustrated by \cite{Lin13}, PSAEM is a competitive alternative to particle-smoothing-based EM algorithms (\eg \citep{SchonWN:2011,OlssonDCM:2008}),
as it enjoys better convergence properties and has a much lower computational cost.
The method
maintains a stochastic approximation of the auxiliary quantity \eqref{eq:Qdef}, $ \widehat Q_k(\btheta) \approx Q(\btheta, \btheta_{k-1})$.
This approximation is updated according to
\begin{align}
  \label{eq:Qsa}
  \widehat Q_k(\btheta) = (1-\gamma_k) \widehat Q_{k-1}(\btheta) + \gamma_k \log p(\y_{0:T}, \x_{0:T}[k] \mid \btheta).
\end{align}
Here, $\{ \gamma_k \}_{k\geq0}$ is a sequence of step sizes, satisfying the usual stochastic approximation
conditions: $\sum_k \gamma_k = \infty$ and $\sum_k \gamma_k^2 < \infty$. A typical choice is to take
$\gamma_k = k^{-p}$ with $p \in \,]0.5, 1]$, where a smaller value of $p$ gives a more rapid convergence at the cost of higher variance.
In the vanilla SAEM algorithm, $\x_{0:T}[k]$ is a draw from the smoothing distribution $p(\x_{0:T} \mid \y_{0:T}, \btheta_{k-1})$.
In this setting, \cite{DeyLavMou99} show that using the stochastic approximation \eqref{eq:Qsa} instead of \eqref{eq:Qdef} in the EM algorithm
results in a valid method, \ie $\{\btheta_k\}_{k\geq0}$ will still converge to a maximizer of
$p(\y_{0:\T} \mid \btheta)$.

The PSAEM algorithm is an extension of SAEM, which is useful when it is not possible to sample
directly from the joint smoothing distribution. This is indeed the case in our setting.
Instead of sampling from the smoothing distribution, the sample trajectory $\x_{0:T}[k]$ in \eqref{eq:Qsa} may be
drawn from an ergodic Markov kernel, leaving the smoothing
distribution invariant. Under suitable conditions on the kernel, this will not violate the validity of SAEM; see \citep{AndrieuV:2011,AndrieuMP:2005}.

In PSAEM, this Markov kernel on the space of trajectories, denoted as ${\kernelPGAS(\x_{0:\T}^\star \mid \x_{0:\T}^\prime)}$, is constructed using PMCMC theory.
In particular, we use the method by \cite{LinJorSch12}, particle Gibbs with ancestor sampling (PGAS).
We have previously used PGAS for Bayesian identification of \gpssm{s} \citep{FriLinSchRas13}.
PGAS is a sequential Monte Carlo method, akin to a standard particle filter (see \eg \citep{DoucetJ:2011,Gustafsson:2010a}),
but with the difference that one particle at each time point is specified \emph{a priori}.
These reference states, denoted as $\x_{0:\T}^\prime$, can be thought of as guiding the particles of the particle
filter to the ``correct'' regions of the state-space. More formally, as shown by \cite{LinJorSch12},
PGAS defines a Markov kernel which leaves the joint smoothing distribution invariant, \ie for any $\btheta$,
\begin{align}
  \label{eq:invariance}
  \int \kernelPGAS(\x_{0:\T}^\star \mid \x_{0:\T}^\prime) p(\x_{0:T}^\prime \mid \y_{0:T}, \btheta)\myd\x_{0:T}^\prime
  = p(\x_{0:T}^\star \mid \y_{0:T}, \btheta).
\end{align}
The PGAS kernel is indexed by $N$, which is the number of particles used in the underlying particle filter.
Note in particular that the desired property \eqref{eq:invariance} holds for any ${N \geq 1}$,
\ie the number of particles only affects the mixing of the Markov kernel. A larger $N$ implies
faster mixing, which in turn results in better approximations of the auxiliary quantity \eqref{eq:Qsa}.
However, it has been experienced in practice that the correlation between consecutive trajectories
drops of quickly as $N$ increases \citep{LinJorSch12,LindstenS:2013},
and for many models a moderate $N$ (\eg in the range 5--20) is enough to get a rapidly mixing kernel.
We refer to \citep{Lin13,LinJorSch12} for details.
We conclude by noting that it is possible to generate a sample $\x_{0:\T}[k] \sim \kernelPGAS[{\btheta[k-1]}](\,\cdot \mid \x_{0:\T}[k-1])$
by running a particle-filter-like algorithm. This method is given as Algorithm~1 in \citep{LinJorSch12} and is described specifically
for \gpssm{s} in Section~3 of \citep{FriLinSchRas13}.

Next, we address the M-step of the EM algorithm. Maximizing the quantity \eqref{eq:Qsa}
will typically not be possible in closed form. Instead, we make use
of a numerical optimization routine implementing a quasi-Newton method (BFGS).
Using \eqref{eq:completedataloglik}, the gradient of the complete data log-likelihood can be written as
\begin{align}
  \nonumber
  &\frac{\partial }{\partial \btheta} \log p(\y_{0:\T},\x_{0:\T} \mid \btheta) =
  \sum_{t=0}^{\T} \frac{\partial }{\partial \btheta} \log p(\y_t \mid \x_t, \btheta)\\
  &\hspace{1em}+  \sum_{t=1}^{\T} \frac{\partial }{\partial \btheta} \log p(\x_t \mid \x_{0:t-1}, \btheta) + \frac{\partial }{\partial \btheta} \log p(\x_0 \mid \btheta),
\end{align}
where the individual terms can be computed using \eqref{eq:likelihood} and \eqref{eq:allsequential}, respectively.
The resulting PSAEM algorithm for learning of \gpssm{s} is summarized in Algorithm~\ref{alg:psaem}.

\begin{algorithm}
  \caption{PSAEM for \gpssm{s}}
  \label{alg:psaem}
  \begin{enumerate}
  \item Set $\btheta_0$ and $\x_{0:\T}[0]$ arbitrarily. Set $\widehat Q_0(\btheta) \equiv 0$.
  \item For $k \geq 1$:
    \begin{enumerate}
    \item Simulate $\x_{0:\T}[k] \sim \kernelPGAS[{\btheta[k-1]}](\,\cdot \mid \x_{0:\T}[k-1])$ (run Algorithm~1 in \citep{LinJorSch12}
      and set $\x_{0:\T}[k]$ to one of the particle trajectories with probabilities given by their importance weights).
    \item Update $\widehat Q_k(\btheta)$ according to \eqref{eq:Qsa}.
    \item Compute $\btheta_k = \argmax_{\btheta} \widehat Q_k(\btheta)$.
    \end{enumerate}
  \end{enumerate}
\end{algorithm}

A particular feature of the proposed approach is that it performs smoothing even when the state transition function is not yet explicitly defined. Once samples from the smoothing distribution have been obtained it is then possible to analytically describe the state transition probability density (see \citep{FriLinSchRas13} for details). This contrasts with the standard procedure where the smoothing distribution is found using a given state transition density.

\section{Experimental Results}

In this section we present the results of applying PSAEM to identify various dynamical systems.

\subsection{Identification of a Linear System}

Although GP-SSMs are particularly suited to nonlinear system identification, we start by illustrating their behavior when identifying the following linear system
\begin{subequations}
\label{eq:linsys}
  \begin{align}
  \label{eq:linstatetransition}
    \x_{t+1} &= 0.8 \, \x_{t} + 3 \, \n{u}_{t} + \n{v}_{t},  &\n{v}_{t} \sim \mathcal{N}(0,1.5), \\
    \n{y}_{t} &= 2 \, \x_{t} + \n{e}_{t}, &\n{e}_{t} \sim \mathcal{N}(0,1.5),
  \end{align}
\end{subequations}
excited by a periodic input. The GP-SSM can model this linear system by using a linear covariance function for the GP. This covariance function imposes, in a somehow indirect fashion, that the state-transition function in~\eqref{eq:linstatetransition} must be linear. A GP-SSM with linear covariance function is formally equivalent to a linear state-space model where a Gaussian prior is placed over the, unknown to us, parameters ($A=0.8$ and $B=3$) \citep[Section 2.1]{RasWil06}. The hyper-parameters of the covariance function are equivalent to the variances of a zero-mean prior over $A$ and $B$. Therefore, the application of PSAEM to this particular GP-SSM can be interpreted as finding the hyper-parameters of a Gaussian prior over the parameters of the linear model that maximize the likelihood of the observed data whilst marginalizing over $A$ and $B$. In addition, the likelihood will be simultaneously optimized with respect to the process noise and measurement noise variances ($q$ and $r$ respectively).

\begin{figure}[t!]
\centering
\includegraphics[width=0.9\linewidth]{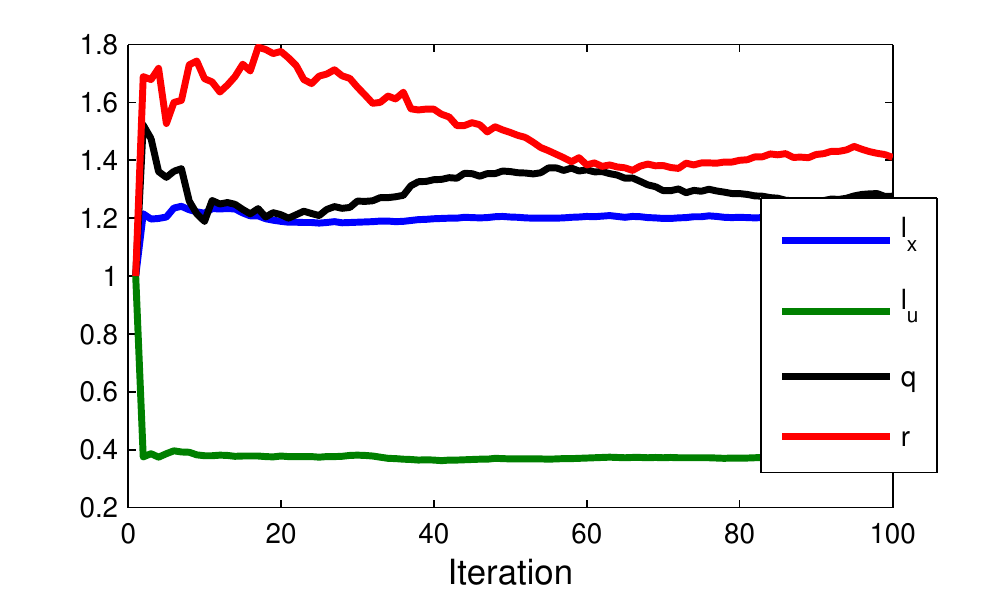}
\caption{Convergence of parameters when learning a linear system using a linear covariance function.}
\label{fig:lin_covlin_converg}
\end{figure}

\begin{figure}[t!]
\centering
\includegraphics[width=1.08\linewidth]{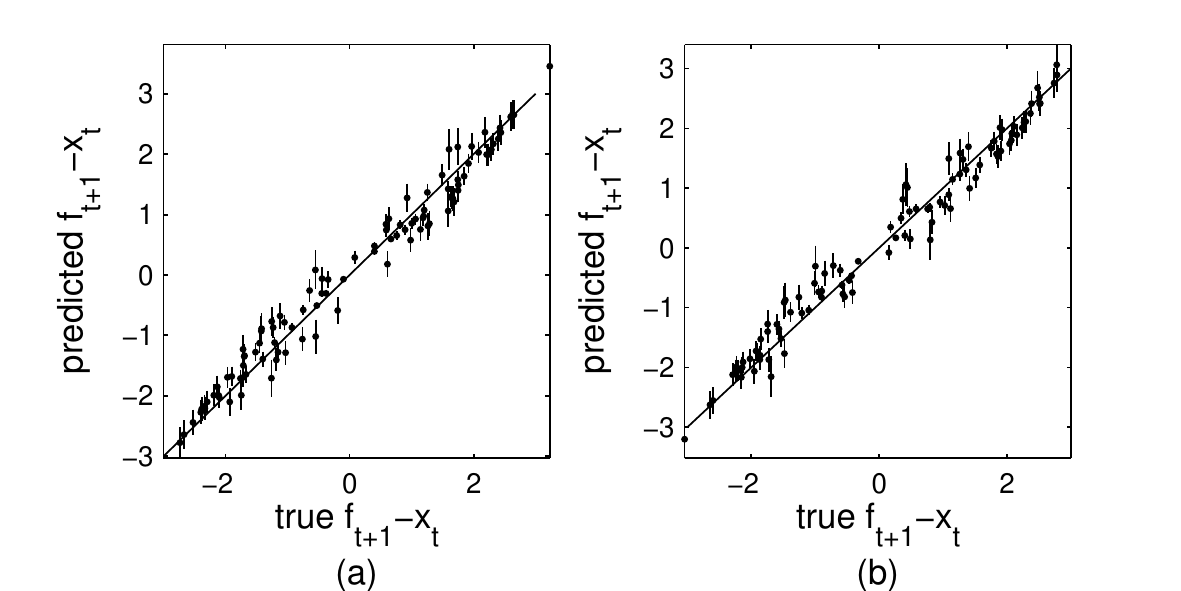}
\caption{Linear dynamical system learned using a GP-SSM with linear covariance function. Predictions (a) on training data, and (b)  on test data (see text for more details).}
\label{fig:testlinearcovlinear}
\end{figure}

Figure~\ref{fig:lin_covlin_converg} shows the convergence of the GP hyper-parameters ($l_x$ and $l_u$) and noise parameters with respect to the PSAEM iteration. In order to judge the quality of the learned GP-SSM we evaluate its predictive performance on the data set used for learning (training set) and on an independent data set generated from the same dynamical system (test set). The GP-SSM can make probabilistic predictions which report the uncertainty arising from the fact that only a finite amount of data is observed. 

Figure~\ref{fig:testlinearcovlinear} displays the predicted value of $\n{f}_{t+1} - \x_t$ versus the true value. Recall that $\n{f}_{t+1} - \x_t$ is equivalent to the step taken by the state in one single transition before process noise is added: $f(\x_t,\n{u}_t) - \x_t$. One standard deviation error bars from the predictive distribution have also been plotted. Perfect predictions would lie on the unit slope line. We note that although the predictions are not perfect, error-bars tend to be large in predictions that are far from the true value and narrower for predictions that are closer to the truth. This is the desired outcome since the goal of the GP-SSM is to represent the uncertainty in its predictions.

\begin{figure}[t!]
\centering
\includegraphics[width=0.9\linewidth]{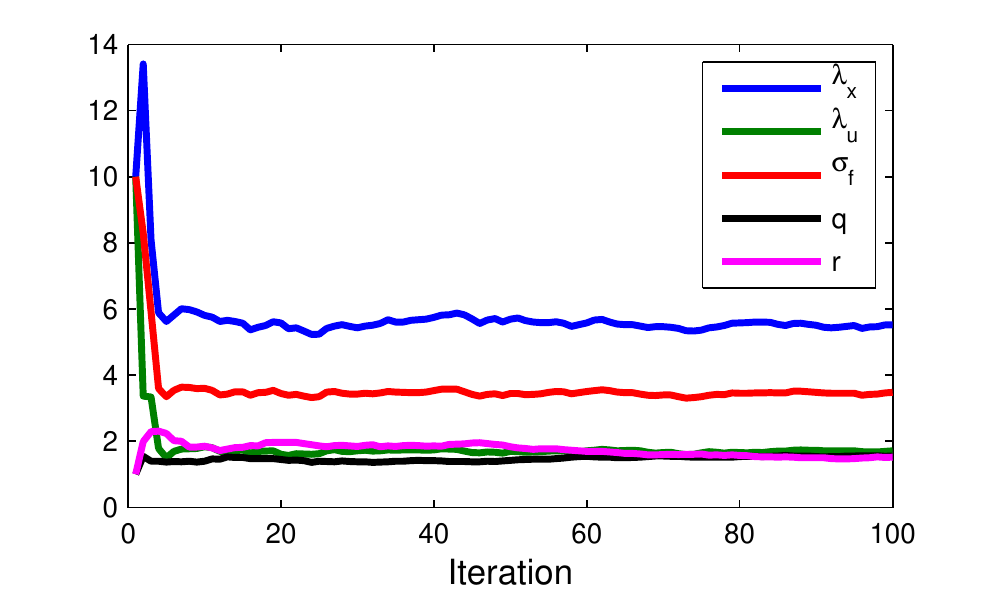}
\caption{Convergence of parameters when learning a linear system using a squared exponential covariance function.}
\label{fig:lin_covse_converg}
\end{figure}

\begin{figure}[t!]
\centering
\includegraphics[width=1.08\linewidth]{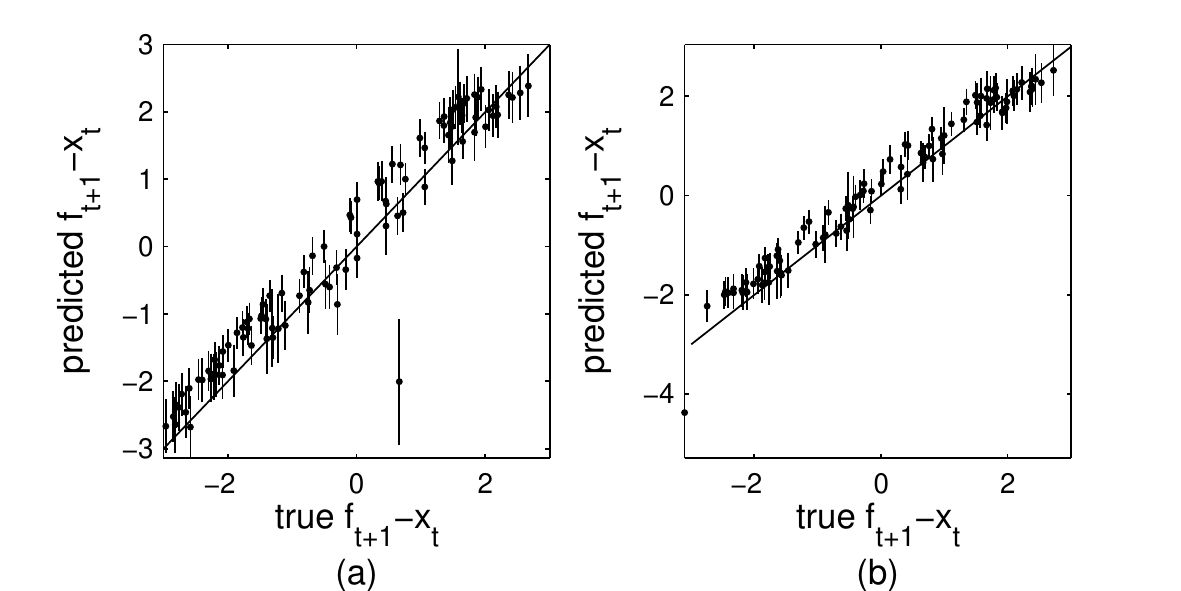}
\caption{Linear dynamical system learned using a GP-SSM with squared exponential covariance function. Predictions (a) on training data, and (b)  on test data.}
\label{fig:testlinearcovsqexp}
\end{figure}

We now move into a scenario in which the data is still generated by the linear dynamical system in~\eqref{eq:linsys} but we pretend that we are not aware of its linearity. In this case, a covariance function able to model nonlinear transition functions is a judicious choice. We use the squared exponential covariance function which imposes the assumption that the state transition function is smooth and infinitely differentiable \citep{RasWil06}. Figure~\ref{fig:lin_covse_converg} shows, for a PSAEM run, the convergence of the covariance function hyper-parameters (length-scales $\lambda_x$ and $\lambda_u$ and signal variance $\sigma_f$) and also the convergence of the noise parameters.

The predictive performance on training data and independent test data is presented in Figure~\ref{fig:testlinearcovsqexp}. Interestingly, in the panel corresponding to training data (a), there is particularly poor prediction that largely underestimates the value of the state transition. However, the variance for this prediction is very high which indicates that the identified model has little confidence in it. In this particular case, the mean of the prediction is 2.5 standard deviations away from the true value of the state transition.


\subsection{Identification of a Nonlinear System}

\begin{figure}[t!]
\centering
\includegraphics[width=1\linewidth]{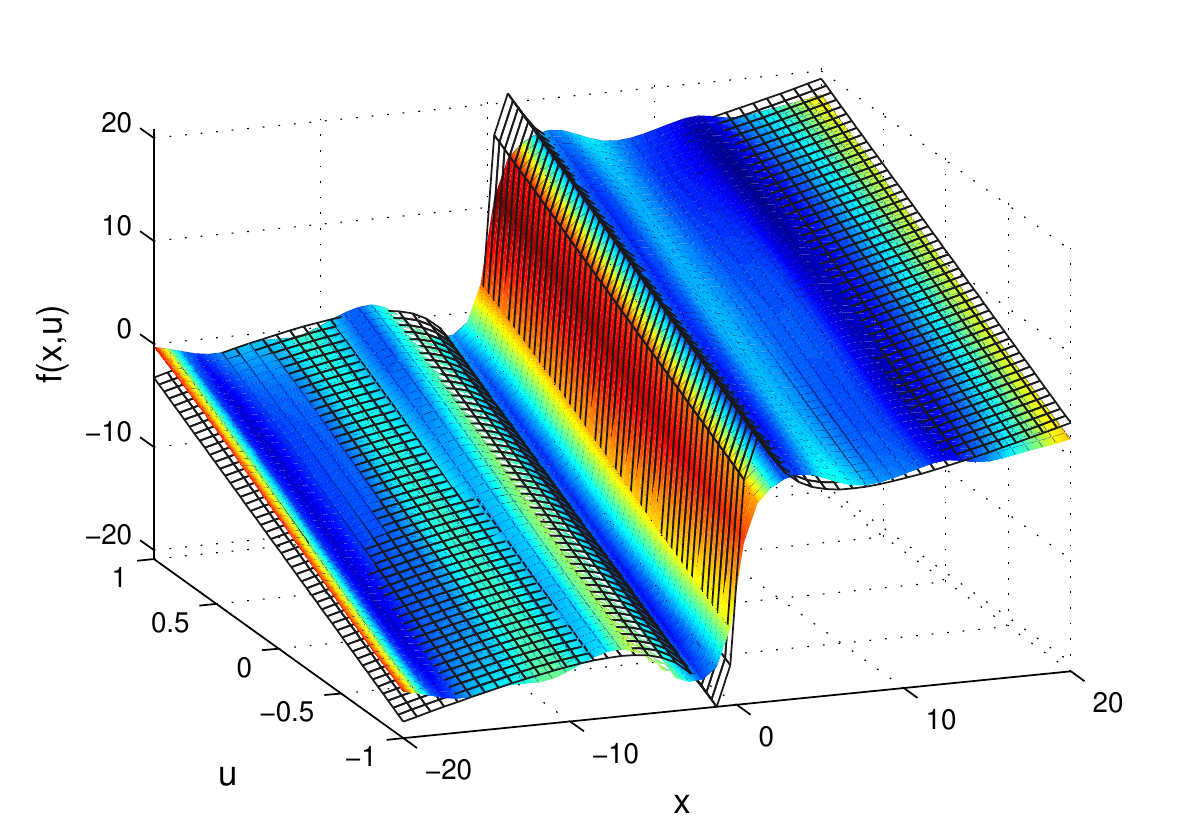}
\caption{Nonlinear dynamical system with one state and one input. The black mesh represents the ground truth dynamics function and the colored surface is the mean of the identified function. Color is proportional to the standard deviation of the identified function  (red represents high uncertainty and blue low uncertainty).}
\label{fig:3d}
\end{figure}

\begin{figure}[t!]
\centering
\includegraphics[width=1.08\linewidth]{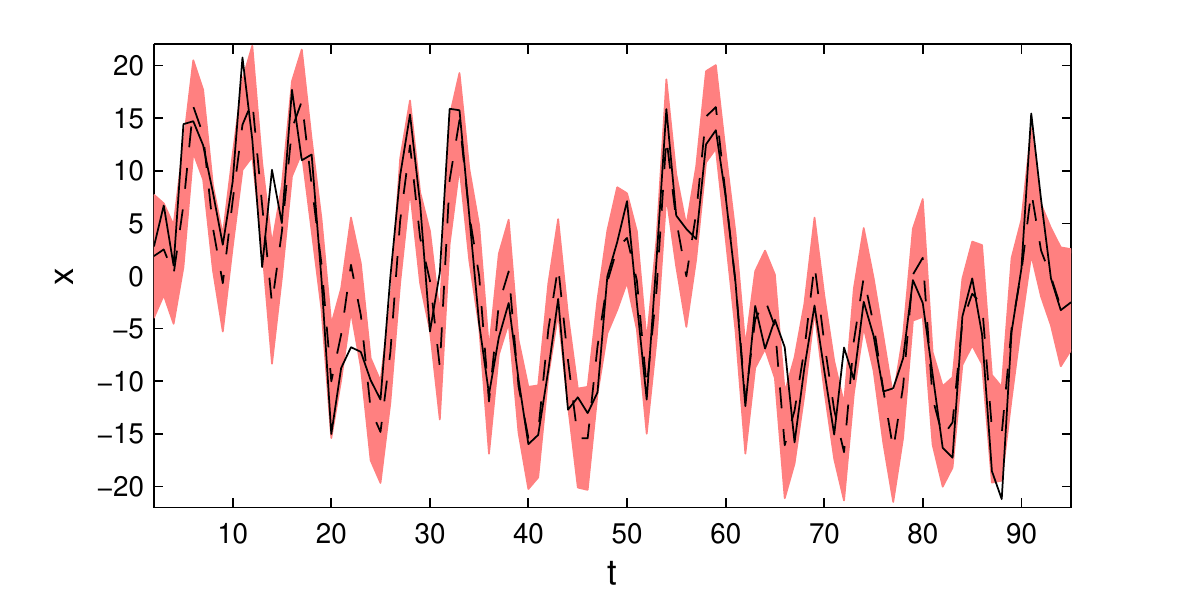}
\caption{State trajectory from a test data set (solid black line). One step ahead predictions made with the identified model are depicted by a dashed line (mean) and a colored interval at $\pm 1$ standard deviation (including process noise).}
\label{fig:nonlinonesteppred}
\end{figure}

GP-SSMs are particularly powerful for nonlinear system identification when it is not possible to create a parametric model of the system based on detailed knowledge about its dynamics. To illustrate this capability of GP-SSMs we consider the nonlinear dynamical system
\begin{subequations}
  \begin{align}
  \label{eq:statenonlin}
    \x_{t+1} &= a \x_{t} + b \frac{\x_t}{1+\x_t^2} + c \n{u}_t + \n{v}_{t}, &\n{v}_{t}
\sim \mathcal{N}(0,q), \\
\label{eq:measurementnonlin}
    \y_{t} &= d \x_t^2 + \n{e}_{t}, &\n{e}_{t} \sim
\mathcal{N}(0,r),
  \end{align}
\end{subequations}
with parameters $(a,b,c,d,q,r) = (0.5,25,8,0.05,10,1)$ and a known input $\n{u}_t = \cos(1.2
(t+1))$. One of the challenging properties of this system is that the quadratic measurement
function~\eqref{eq:measurementnonlin} tends to induce a bimodal distribution in the marginal
smoothing distribution. For instance, if we were to consider only one measurement in isolation and
$r=0$ we would have $\x_t = \pm \sqrt{\frac{\y_t}{d}}$. Moreover, the state transition
function~\eqref{eq:statenonlin} exhibits a very sharp gradient in the $\x_t$ direction at the
origin, but is otherwise parsimonious as $\x_t \rightarrow \pm \infty$.

Again, we pretend that detailed knowledge about the particular form of~\eqref{eq:statenonlin} is not available to us. We select a covariance function that consists of a Mat\'{e}rn covariance function in the $\x$ direction and a squared exponential in the $\n{u}$ direction. The Mat\'{e}rn covariance function imposes less smoothness constraints than the squared exponential \citep{RasWil06} and is therefore more suited to model functions that can have sharp transitions. 

Figure~\ref{fig:3d} shows the true state transition dynamics function (black mesh) and the identified function as a colored surface. Since the identified function from the GP-SSM comes in the form of a probability distribution over functions, the surface is plotted at $\mathbb{E}[\n{f}^* |\x^*, \n{u}^*, \y_{0:\T} ]$ where the symbol~$^*$ denotes test points. The standard deviation of~$\n{f}^*$, which represents our uncertainty about the actual value of the function, is depicted by the color of the surface. Figure~\ref{fig:nonlinonesteppred} shows the one step ahead predictive distributions $p(\x_{t+1}^* | \x_{t}^*, \n{u}_{t}^*, \y_{0:\T})$ on a test data set.





\section{Conclusions}
\gpssm{s} allow for a high degree of flexibility when addressing the nonlinear system identification problem
by making use of Bayesian nonparametric system models.
These models enable the incorporation of high-level assumptions, such as smoothness of the transition function,
while still being able to capture a wide range of nonlinear dynamical functions.
Furthermore, the \gpssm is capable of making probabilistic predictions that can be useful in adaptive control and robotics,
when the control strategy might depend on the uncertainty in the dynamics. Our particle-filter-based maximum likelihood inference of the model hyper-parameters preserves the full nonparametric richness of the model. In addition, marginalization of the dynamical function effectively averages over all possible dynamics consistent with the GP prior and the data, and hence provides a strong safeguard against overfitting.


 
 

 

\bibliographystyle{apalike} 
\bibliography{gp-ssm-psaem,references}

\end{document}